%% file: Main_paper.tex
\documentclass[lettersize,journal]{IEEEtran}
\usepackage{amsmath,amsfonts}
\usepackage{algorithmic}
\usepackage{algorithm}
\usepackage{array}
\usepackage[caption=false,font=normalsize,labelfont=sf,textfont=sf]{subfig}
\usepackage{textcomp}
\usepackage{stfloats}
\usepackage{url}
\usepackage{verbatim}
\usepackage{graphicx}
\usepackage{cite}
\usepackage{amsmath}
\usepackage{amssymb}
\usepackage{booktabs}
\usepackage{multirow}
\usepackage{bbding}
\usepackage{makecell}
\usepackage{amsfonts}
\usepackage{comment}
\usepackage{xcolor}
\usepackage{epstopdf}

\usepackage{caption}
\usepackage{hyperref}
\hypersetup{hidelinks} 
\newcommand{\fengx}[1]{\textcolor[rgb]{1,0,1} {#1}}

\begin{document}

\title{Global-Local Stepwise Generative Network for Ultra High-Resolution Image Restoration}

\author{Xin Feng, Haobo Ji, Wenjie Pei, Fanglin Chen,~\IEEEmembership{Member,~IEEE}, and Guangming Lu,~\IEEEmembership{Senior Member,~IEEE}

\thanks{X. Feng, H. Ji, W. Pei, F. Chen and G. Lu are with the Department of Computer Science, Harbin Institute of Technology at Shenzhen, Shenzhen 518057, China. E-mail: \href{mailto:fengx_hit@outlook.com}{fengx\_hit@outlook.com}; \href{mailto:20s151172@stu.hit.edu.cn}{20s151172@stu.hit.edu.cn}; \href{mailto:wenjiecoder@outlook.com}{wenjiecoder@outlook.com}; \href{mailto:chenfanglin@hit.edu.cn}{chenfanglin@hit.edu.cn}; \href{mailto:luguangm@hit.edu.cn}{luguangm@hit.edu.cn}. 

This work has been submitted to the IEEE for possible publication. Copyright may be transferred without notice, after which this version may no longer be accessible.
}}




\maketitle

\input{0.abstract}


\section{Introduction}
\label{sec:intro}
\input{1.introduction}

\section{Related Work}
\label{sec:relatedwork}
\input{2.relatedwork}

\section{Global-Local Stepwise Generative Network}
\label{sec:method}
\input{3.method}

\section{UHR4K Dataset}
\label{sec:data}
\input{4.0.dataset}

\section{Experiments}
\label{sec:exp}
\input{4.experiments}

\section{Conclusion}
\label{sec:con}
\input{5.conclusion}


\bibliographystyle{IEEEtran}
\bibliography{egbib}












\newpage

 





\end{document}

%% file: 0.abstract.tex
\begin{abstract}
While the research on image background restoration from regular size of degraded images has achieved remarkable progress, restoring ultra high-resolution (e.g., 4K) images remains an extremely challenging task due to the explosion of computational complexity and memory usage, as well as the deficiency of annotated data. In this paper we present a novel model for ultra high-resolution image restoration, referred to as the Global-Local Stepwise Generative Network (\emph{GLSGN}), which employs a stepwise restoring strategy involving four restoring pathways: three local pathways and one global pathway. The local pathways focus on conducting image restoration in a fine-grained manner over local but high-resolution image patches, while the global pathway performs image restoration coarsely on the scale-down but intact image to provide cues for the local pathways in a global view including semantics and noise patterns. 
To smooth the mutual collaboration between these four pathways, our \emph{GLSGN} is designed to ensure the inter-pathway consistency in four aspects in terms of low-level content, perceptual attention, restoring intensity and high-level semantics, respectively.
As another major contribution of this work, we also introduce the first ultra high-resolution dataset to date for both reflection removal and rain streak removal, comprising 4,670 real-world and synthetic images.
Extensive experiments across three typical tasks for image background restoration, including image reflection removal, image rain streak removal and image dehazing, show that our \emph{GLSGN} consistently outperforms state-of-the-art methods.
Code is released at \fengx{\url{https://github.com/F-Frida/GLSGN}}.

\end{abstract}

%% file: 1.introduction.tex
\IEEEPARstart{I}{mage} restoration is a fundamental research topic in computer vision due to its extensive applications including image reflection removal~\cite{li2020single,wei2019single}, image rain streak removal~\cite{chen2021robust,wang2021rain,jiang2021multi} and image dehazing~\cite{zhou2022fsad,wu2021contrastive,han2018review}. The methods based on deep generative framework~\cite{zou2020deep,feng2021deep,yang2016single,zamir2021multi} have brought about great progress in image restoration from regular size of degraded images. Nevertheless, 
performing image restoration on ultra high-resolution (e.g., 4K, 3840$\times$2160) images remains a challenging task, which demands quadratically more computational power and memory storage than the routine image restoration on regular size of images. Moreover, higher-resolution degraded images involve more sophisticated noise patterns including more scales of noise and more diverse noise patterns. Thus, it is impractical to directly apply existing methods for routine image restoration to it. Another important factor that hinders the research on image restoration for ultra high-resolution images is the deficiency of ultra high-resolution benchmark datasets for image restoration. In this work, we aim to address both these two limitations in terms of algorithm and data respectively.

A straightforward way to deal with high-resolution images is to employ the global-local strategy~\cite{chen2019GLNET,zhang2020local,zamir2021multi,zhang2021global,zhou2019global}, which operates on each divided local region and then fuses the local results based on the global view. A crux when applying such global-local processing strategy is how to model the global-local interactions to ensure the consistency of the whole restored image.
In this work we aim to adapt such global-local strategy to image restoration for ultra high-resolution images while focusing on modeling the global-local consistency. Specifically, we propose the Global-Local Stepwise Generative Network (\emph{GLSGN}), which performs ultra high-resolution image restoration in a stepwise manner employing two types of restoring pathways: three local pathways and one global pathway. The local pathways focus on conducting fine-grained image restoration over local but high-resolution patches, while the global pathway is designed to perform image restoration coarsely on scale-down but intact images to capture global information including noise patterns and high-level semantics. Referring to such global cues provided by the global pathway, three local pathways are able to restore images successively with increasing receptive fields to achieve the clean background progressively.

To model the interactions between different pathways and thus ensure the inter-pathway consistency of our \emph{GLSGN}, we model the inter-pathway interactions during the image restoration process, thereby smoothing the mutual collaboration between the four pathways. To be specific, we ensure four types of consistency between pathways: 1) low-level content consistency by iterative decoding between pathways, 2) perceptual attention consistency by coordinating the attention distributions of different pathways, 3) the consistency of restoring intensity by inter-patch normalization and 4) high-level semantics consistency using Laplacian Pyramid for background synthesis. 

To address the limitation of data deficiency for ultra high-resolution image reflection removal and rain streak removal, we collect a large dataset, named UHR4K, comprising totally 4,670 real-world and synthetic images. In particular, it's the first large scale benchmark for evaluating performance of image reflection removal and rain streak removal in 4K resolution, and includes various noise intensities and sufficient scenes for learning to restore ultra high-resolution background image.
To conclude, we have following contributions:
\begin{itemize}
    \item We propose the novel \emph{GLSGN} model for ultra high-resolution image restoration, 
     which employs three local pathways and one global pathway to perform image restoration collaboratively. In particular, four types of inter-pathway consistency are particularly modeled to ensure the global-local consistency in the obtained high-resolution image.
    \item We construct a large and high-quality 4K dataset named \emph{UHR4K}, which is the first for image restoration tasks including image reflection removal and image rain streak removal.
    \item Extensive experiments across three image restoration tasks including reflection removal, rain streak removal and image dehazing, demonstrate that our proposed \emph{GLSGN} outperforms other state-of-the-art methods consistently. 
\end{itemize}

%% file: 2.relatedwork.tex
\noindent\textbf{Generic methods for image restoration.}
Benefiting from the performance improvement by the attention mechanism~\cite{vaswani2017attention}, some researchers~\cite{zhang2019rnan,liu2018non,mei2020pyramid, tian2021asymmetric,feng2021deep, wang2021uformer} proposed generic frameworks for image restoration, which learn to precisely model the noise patterns by attention maps.
For instance, \cite{zhang2019rnan} and \cite{liu2018non} incorporated non-local attention modules into deep networks to capture long-term dependencies between pixels and pay attention to the challenging noise patterns.
Considering the cross-scale dependency, Mei \emph{et al.}~\cite{mei2020pyramid} proposed a pyramid attention network, which learns the self-similarity of noise patterns by a non-local pyramid module.
Later, Some researchers~\cite{wang2021uformer,ji2021u2,chen2021pre} attempt to employ the self-attention mechanism to improve the performance of image restoration.
Wang \emph{et al.}~\cite{wang2021uformer} first exploited a U-shaped transformer for image restoration by embedding the self-attention block into a U-shaped structure.
Based on this work, Ji \emph{et al.}~\cite{ji2021u2} further designed a nested transformer framework to improve the quality of restored images.
To deal with multiple image restoration tasks, Chen \emph{et al.}~\cite{chen2021pre} developed a pre-trained image processing transformer to model diverse noise patterns.

Another category of generic methods~\cite{pan2020physics,zou2020deep,pan2021exploiting} for image restoration employs the generative adversarial net (GAN)~\cite{goodfellow2014generative} to enhance the quality of restored results.
By virtue of GAN's ability to learn data distribution, Zou \emph{et al.}~\cite{zou2020deep} designed a unified framework for background image restoration and performed adversarial learning to separate superimposed images.
Additionally, Pan \emph{et al.}~\cite{pan2021exploiting} manifested the effectiveness of generative priors by pre-trained GANs in various image restoration tasks.
The last category of methods designs multi-stage frameworks to enlarge the receptive field of modeling noise patterns.
Suin \emph{et al.}~\cite{suin2020spatially} proposed a lightweight framework in a progressive manner, aiming to model different blurring degradations.
Recently, Zamir \emph{et al.}~\cite{zamir2021multi} proposed a generic multi-stage framework for image restoration, which follows the progressive manner of restoring increasingly larger image patches for modeling diverse noise patterns. 
In summary, present generic methods for image restoration perform well on regular-resolution images, but are heavily limited on the performance for high-resolution images because of the explosion of computational complexity and memory usage.
Besides, due to the increase of pixels, generic methods are also hard to model intact noise patterns in high-resolution images.

\noindent\textbf{Global-local consistency-based methods for image restoration.}
Learning global-local consistency is a classic modeling strategy, and has been validated its effectiveness in various computer vision tasks.
A typical way of learning global-local consistency is inspired by the inception series~\cite{szegedy2015going}, which extract features by different size of convolutional kernels, and then aggregate features together.
Inspired by the image pyramid, Xu \emph{et al.}~\cite{xu2021edpn} further introduced a deep pyramid network for blurry image restoration from multiple levels of degradation, which exploits the self- and cross-scale similarities in the degraded image.
Another way of learning global-local consistency~\cite{zhang2019deep,suin2020spatially,zhang2020local,zamir2021multi} first employs different networks to process the global image and local patches separately, and then aggregates them to reconstruct the final result.
Zhang \emph{et al.}~\cite{zhang2019deep} presented a deep hierarchical multi-patch network, which restores blurry images in a coarse-to-fine manner.
Based on this work, Suin \emph{et al.}~\cite{suin2020spatially} designed a content-aware global-local filtering module that significantly improves performance by considering global dependencies and dynamically exploiting neighboring pixel information.
Similarly, Zhang \emph{et al.}~\cite{zhang2020local} proposed a novel local–global dual-stream network that adaptively captures local and long-range information for remote sensing images.
Although global-local consistency contributes to modeling intact noise patterns, it has not yet been fully exploited in high-resolution image processing.

\noindent\textbf{Ad hoc methods for ultra high-resolution image restoration.}
Unlike aforementioned methods which mainly process regular-resolution images, methods for ultra high-resolution image restoration need to tackle the challenge resulting from the explosion of computational complexity and memory usage.
A straightforward way~\cite{yi2020contextual,chen2019GLNET,lin2021real,esser2021taming,zhang2021escnet} of processing high-resolution images is to design a lightweight framework.
Yi \emph{et al.}~\cite{yi2020contextual} introduced a novel contextual residual aggregation mechanism by learning the change of image resolution for ultra high-resolution image inpainting. 
By dividing high-resolution images into local patches, Chen \emph{et al.}~\cite{chen2019GLNET} significantly reduced the computational cost.
Later, Zhang \emph{et al.}~\cite{zhang2021escnet} proposed an end-to-end superpixel-enhanced change detection network for remote sensing image processing, which combines differentiable superpixel segmentation and a deep convolutional neural network.
Recently, for ultra high-resolution image dehazing, Zheng \emph{et al.}~\cite{zheng2021ultra} presented a lightweight framework by introducing the bilateral filtering, which is able to restore 4K level haze-free images.
Despite the fact that these methods achieve improvement, they are primarily designed for specific tasks. It is hard to model various noise types and work as generic methods for ultra high-resolution image restoration. 
Thus, in this paper, we present a generic framework for ultra high-resolution image restoration, which is adaptive for various restoration tasks like image reflection removal, image deraining, and image dehazing.
Our proposed framework achieves much better performance in terms of computational cost and restoration quality than previous methods.




%% file: 3.method.tex
We present the Global-Local Stepwise Generative Network (\emph{GLSGN}) for ultra high-resolution image restoration, which is illustrated in Figure~\ref{fig1:frame}. 
The proposed \emph{GLSGN} employs stepwise restoring strategy involving four restoring pathways: three local pathways and one global pathway. The local pathways focus on conducting image restoration over local but high-resolution patches, while the global pathway performs image restoration coarsely over a scale-down but intact image to provide cues in a global view including semantics and noise patterns for local pathways. All four pathways follow the encoder-decoder generative framework. 

We will first introduce the whole framework of our model briefly, and then elaborate on how the four pathways of our model perform image restoration collaboratively and achieve the global-local consistency between each other.

\subsection{Global-Local Stepwise Generative Framework}
As shown in Figure~\ref{fig1:frame}, the three local pathways of our \emph{GLSGN} operate on three different sizes of images with decreasing resolutions and perform image restoration in a stepwise manner. The first pathway operates on each of sixteen local patches partitioned equally from the full-resolution image ($4K$) and the decoded feature maps are fed into the second local pathways, which performs restoration on patches from scaled-down images with a quarter of original resolution ($2K$). The third and the last pathway takes the input image with $1/16$ of initial size as input and combines the decoded features from the second pathway to perform image restoration. In this way, these three local pathways restore the background successively and achieve progressively cleaner background image. Meanwhile, the global pathway performs image restoration on the whole input image but with only $1/16$ of initial resolution to obtain a rough restoration result efficiently to provide cues in a global view for the three local pathways.

Formally, the $i$-th local pathway $\mathcal{S}_i$ of our \emph{GLSGN} performs image restoration by:
\begin{equation}
    \hat{\mathbf{B}}_i = \mathcal{S}_i(\mathbf{I}_i,\textbf{F}_{i-1},\textbf{F}_{\text{g}}),
\label{eqn:content_consis}
\end{equation}
where $\hat{\mathbf{B}}_i$ is the restored background image from the input patches $\mathbf{I}_i$, and $\mathbf{F}_{i-1}$ and $\mathbf{F}_{\text{g}}$ are the decoded features from the previous pathway $\mathcal{S}_{i-1}$ and the global pathway $\mathcal{S}_{g}$ respectively ($\mathcal{S}_{1}$ have no previous pathway and only receives feedback from $\mathcal{S}_{g}$). 
All four pathways share the same structure of encoders but with independent parameters. Note that the 
input patches for different pathways have different sizes of receptive fields corresponding to the initial intact image: the patches from higher resolution images correspond to smaller receptive field in the initial image.

\subsection{Global-Local Consistency between Pathways for Collaborative Image Restoration}
Since the input patches of different pathways have different resolutions and different sizes of receptive fields, the crux of our \emph{GLSGN} lies in the global-local consistency between four pathways to perform  image restoration collaboratively. Our \emph{GLSGN} is designed to ensure the global-local consistency in four aspects: 1) restored low-level content, 2) perceptual attention, 3) restoring intensity and 4) high-level semantics.

\begin{figure}[!tp]
    \centering
    \begin{minipage}[b]{1\linewidth}
    \includegraphics[width=\linewidth]{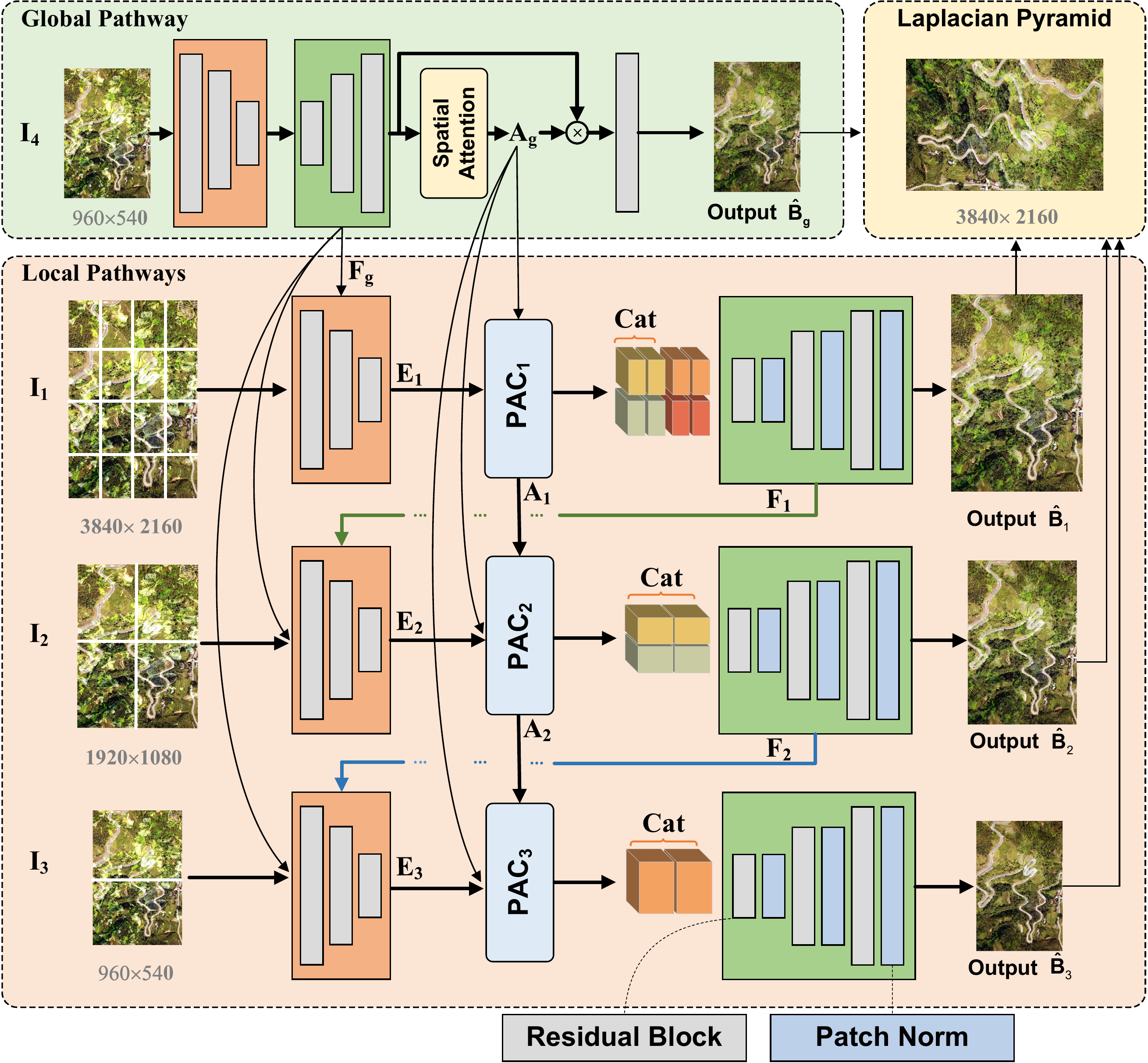}
    \end{minipage}
    \caption{Proposed global-local stepwise generative network for ultra high-resolution image restoration. `Cat' denotes concatenation of patches to intact feature maps, and $\otimes$ is the element-wise multiplication.
    }
    \label{fig1:frame}
\end{figure}

\smallskip\noindent\textbf{Consistency of restored low-level content by iterative decoding.}
To guarantee the consistency of low-level content restored by different pathways, we feed the decoded features of the previous pathway and the global pathway into the current pathway by concatenating them with the corresponding encoded layer of current pathway, as indicated in Equation~\ref{eqn:content_consis} and Figure~\ref{fig1:frame}. As a result, the decoding process of current pathway is performed referring to the global cues from the global pathways and fine-grained cues from the previous pathways.

\begin{figure}[!t]
    \centering
    \includegraphics[width=\linewidth]{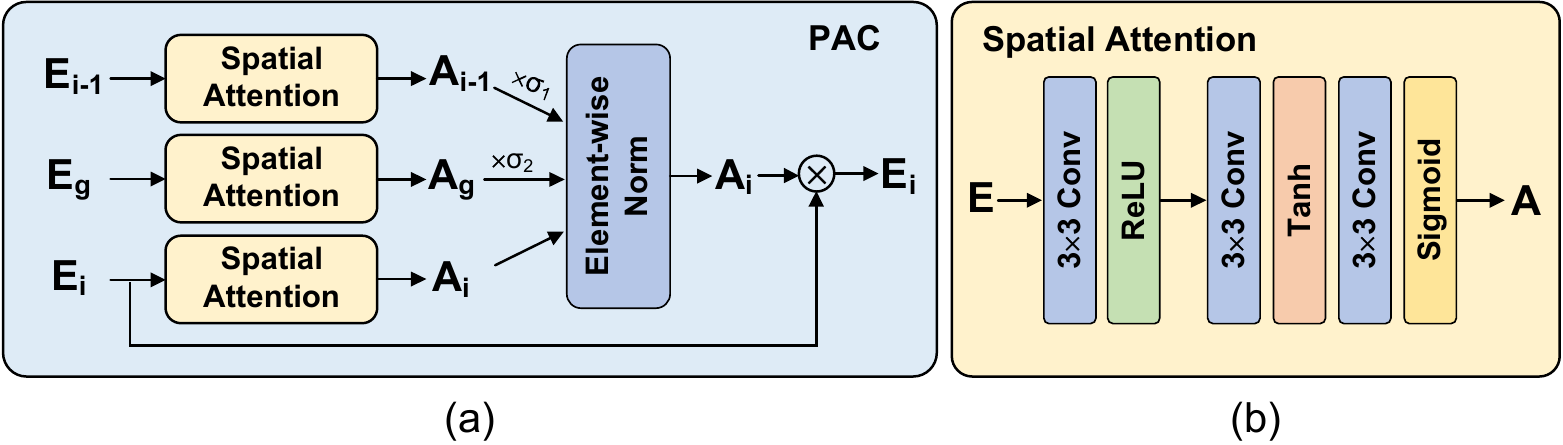}
    \vspace{-8pt}
    \caption{(a) The structure of perceptual attention consistency (PAC) mechanism. (b) The structure of spatial attention.
    }
    \label{fig2:PAC}
\end{figure}

\smallskip\noindent\textbf{Perceptual Attention Consistency.}
Intuitively, the regions with heavy noise in an input image tend to obtain more attention than the regions with light noise during background restoration. The noise patterns in different scales of a same original image typically remain the same, thus four pathways in our \emph{GLSGN} should share the consistent perceptual attention during background restoration. To this end, we propose the Perceptual Attention Consistency (\emph{PAC}) mechanism as shown in Figure~\ref{fig2:PAC}(a). Specifically, for each pathway we employ a spatial attention module~\cite{feng2021deep} shown in Figure~\ref{fig2:PAC}(b) to calculate an attention map from the encoded features, which indicates the spatial perceptual attention for the input image. Then we normalize the perceptual attention of current pathway by fusing the attention maps from the previous pathway and the global pathway:
\begin{equation}
    \mathbf{A}_i = \frac{f^i_\text{att}(\mathbf{E}_i)+\sigma_1\cdot f^{i-1}_\text{att}(\mathbf{E}_{i-1})+\sigma_2\cdot f^g_\text{att}(\mathbf{F}_g)}{1+\sigma_1+\sigma_2},
\end{equation}
where $\mathbf{E}_i$ denotes the encoded features in the $i$-th pathway. Here we use the decoded features of global pathway $\mathbf{F}_g$ to provide a more accurate estimation for attention distribution. $f^i_\text{att}$ denotes the spatial attention module while $\sigma_1$ and $\sigma_2\in$[0, 1] are the hyper-parameters to balance between terms. $\mathbf{A}_i$ is the normalized attention map for $i$-th pathway, which is further used to re-weight the feature maps $\mathbf{E}_i$ by element-wise multiplication $\odot$:
\begin{equation}
    \mathbf{E}_i = \mathbf{A}_i \odot \mathbf{E}_i.
\end{equation}

\smallskip\noindent\textbf{Consistency of restoring intensity by inter-patch normalization.}
\begin{figure}[!t]
    \centering
    \begin{minipage}[b]{0.95\linewidth}
    \includegraphics[width=\linewidth]{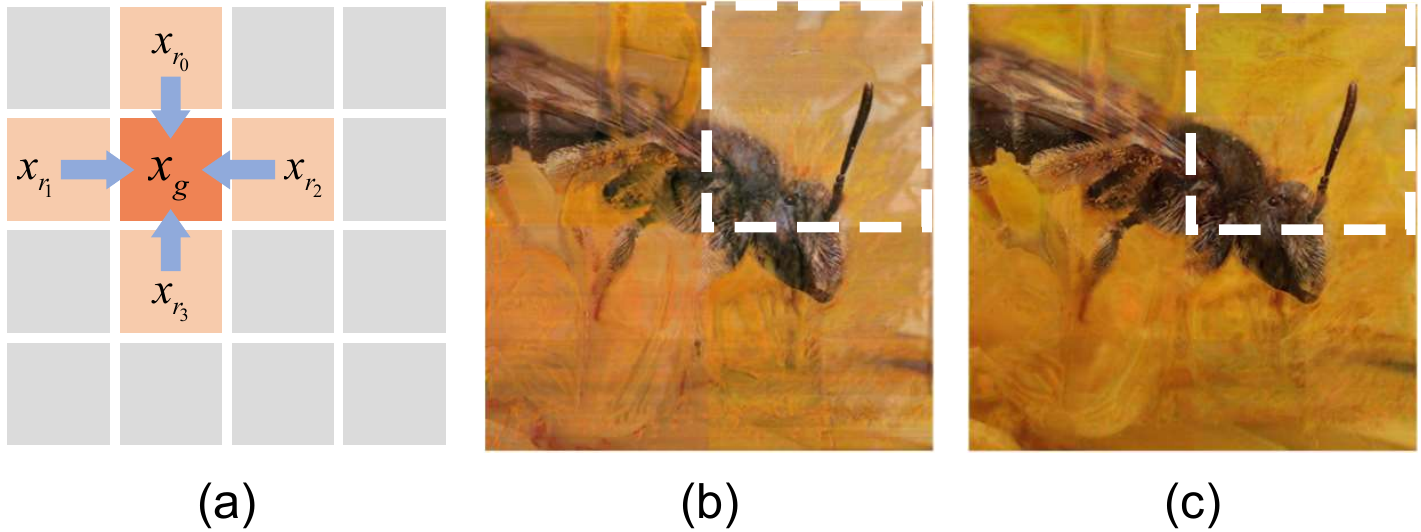}\vspace{4pt}
    \end{minipage}
    \vspace{-10pt}
    \caption{An example of reflection removal for our proposed patch normalization(PN). (a) The visualization of equation~\ref{equ1:PN}. (b) The restored result without patch normalization. (c) The restored result by patch normalization.
    }
    \label{fig4:PN}
\end{figure}
Each local pathway operates on the equally partitioned patches individually during both encoding and decoding. To keep the consistency of restoring intensity between different patches by a same pathway, 
we introduce the patch normalization (PN) that leverages statistics of adjacent patches to regularize the restoration of local patch, which is integrated into each residual block of decoder for each local pathway. Concretely, we first measure the restoring intensity for a patch by calculating the average variation ratios of all pixel values. Then for a patch to be regularized, we calculate its restoring intensity and the average restoring intensity of four adjacent patches $x_r$, and compute the ratio as the regularizing factor. Note that all calculations are based per channel, thus the regularizing factor 
$\mathcal{A}\in\mathbb{R}^{C}$ for patch $x_g$ is calculated as:
\begin{equation}
    \label{equ1:PN}
    \mathcal{A}_{\hat{x}_g}=\frac{\sum\limits^n_{r=1}\sum\limits^{W}_{i=1}\sum\limits^{H}_{j=1}\left|\frac{\hat{x}_{r, \langle i,j \rangle}}{x_{r, \langle i,j \rangle}}\right|}{n\cdot\sum\limits^{W}_{i=1}\sum\limits^{H}_{j=1}\left|\frac{\hat{x}_{g, \langle i,j \rangle}}{x_{g, \langle i,j \rangle}}\right|},
\end{equation}
where ${x}_{g, \langle i,j \rangle},\hat{x}_{g, \langle i,j \rangle}\in\mathbb{R}^{C\times H\times W}$ are the input and output channel-wise pixel of patch-wise feature maps at position $\langle i,j \rangle$ in the same residual block, and $n$ is the number of adjacent patches. In our implementation, we select four adjacent patches around patch $x_g$ as shown in Figure~\ref{fig4:PN}(a) and thus $n$ is 4.
Then the patch
\begin{equation}
    \label{equ2:PN}
    \hat{x}'_g=\mathcal{A}_{\hat{x}_g}\cdot\hat{x}_g+\mathcal{B}(\hat{x}_g),
\end{equation}
where $\mathcal{B}$ learns a bias by additional convolution layer. 
Figure~\ref{fig4:PN}(b) and Figure~\ref{fig4:PN}(c) further illustrate the effect of applying our proposed PN module, and it's easy to observe that our PN modules enable our framework to obtain more consistent background image between different patches.

\begin{figure}[!t]
    \centering
    \begin{minipage}[b]{1\linewidth}
    \includegraphics[width=\linewidth]{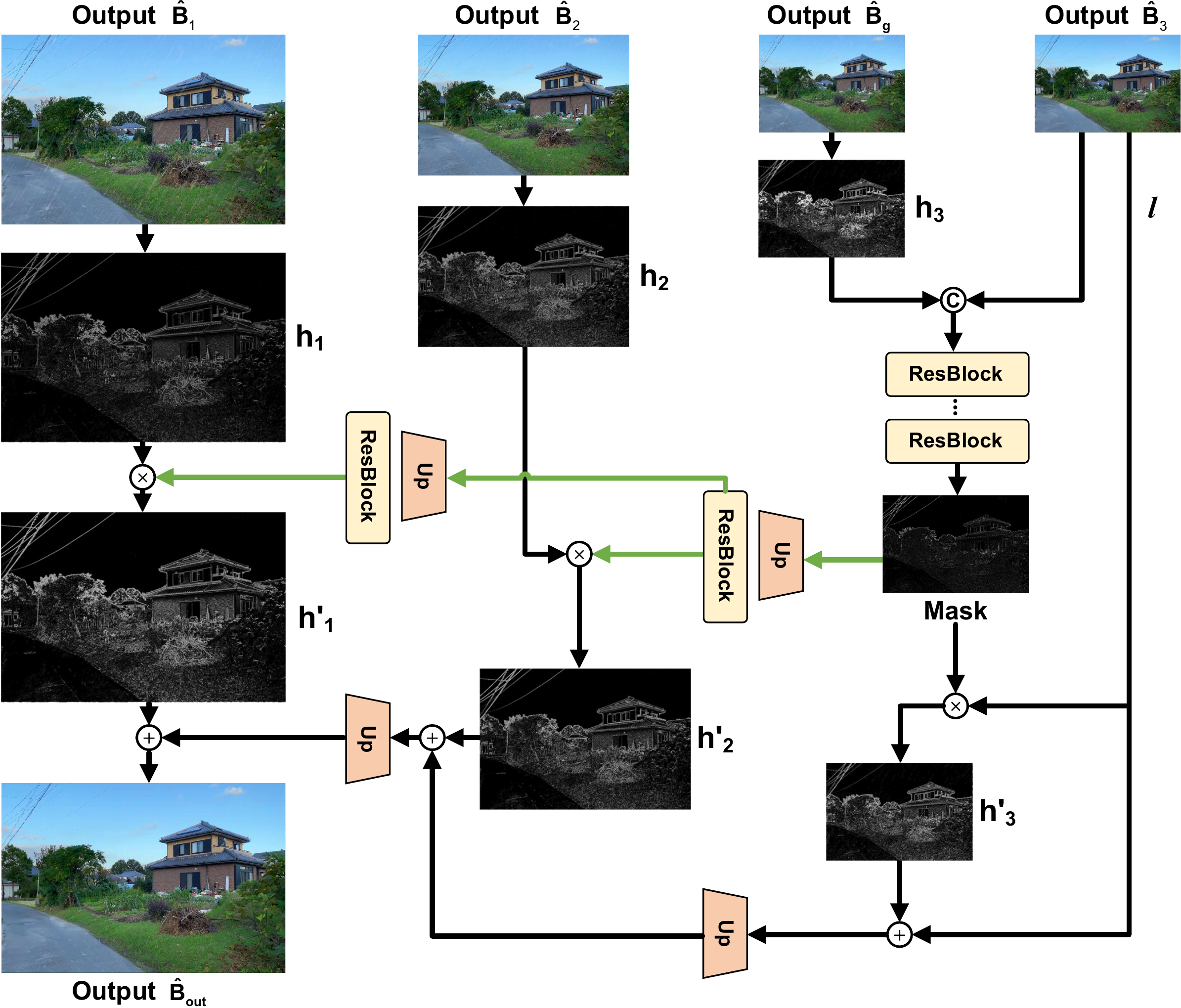}
    \end{minipage}
    \caption{Laplacian Pyramid for reconstructing full resolution ultra high-resolution background image. The green arrow denotes the mask updating steps.
    }
    \label{fig4:laplacian}
\end{figure}
\smallskip\noindent\textbf{Semantic consistency using Laplacian Pyramid for background synthesis.}
Four pathways generate four restored background images with different resolutions. It is crucial to ensure the high-level semantic consistency of these four background images for synthesizing a high-resolution background image from these four background images.
Inspired by Laplacian Pyramid (LP)~\cite{liang2021high,burt1987laplacian} for synthesizing high resolution images, we establish the Laplacian Pyramid based on four restored background images to reconstruct the final high-resolution background image.

\begin{figure*}[!t]
    \centering
    \begin{minipage}[b]{1.0\linewidth}
    \includegraphics[width=\linewidth]{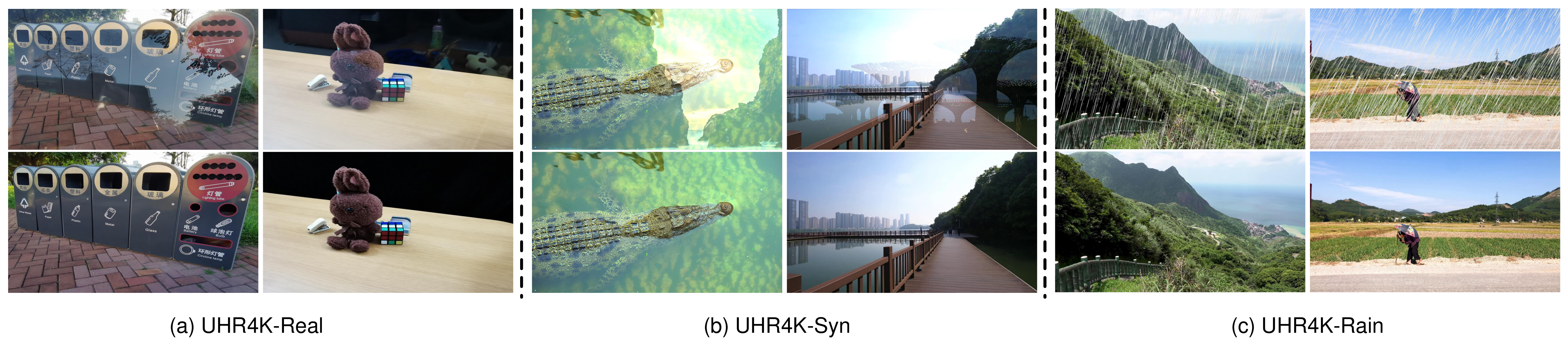}
    \end{minipage}
    \caption{The 4K resolution samples from our UHR4K dataset. Top: images with reflection or rain streak. Bottom: the corresponding groundtruth background images.}
    \label{fig5:dataset}
\end{figure*}
Laplacian pyramid (LP)~\cite{burt1987laplacian}  has a widespread application in image processing, especially image blending~\cite{burt1987laplacian} and super-resolution~\cite{lai2017deep}. The crux of LP is to linearly decouple an image into high-frequency and low-frequency components with different resolutions, from which the original image can be reconstructed invertibly and accurately. Specifically, let $d(\cdot)$ and $u(\cdot)$ be a downsampling operation and an upsampling operator, respectively.
Given an arbitrary image $I_0 \in \mathbb{R}^{H \times W} $, it is first downsampled to obtain a low-frequency prediction $I_1 \in \mathbb{R}^{\frac{H}{2} \times \frac{W}{2}}$, as $I_1=d(I_0)$. To guarantee invertible reconstruction of the original image, the LP records the high-frequency residual $h_0$, as $h_0 = I_0-u(I_1)$. To further reduce image resolution, the LP iteratively performs above operations, obtaining a sequence of low- and high-frequency components. 
Reconstruction from the LP sequence is conducted using the backward recurrence: $I_k = u(I_{k+1}) + h_k$, where $k$ is the number of levels in the pyramid.


Following the above steps, we establish the LP based on four restored background images to reconstruct the final high-resolution background image, as illustrated in Figure~\ref{fig4:laplacian}. We first extract the high-frequency components [$h_1$, $h_2$, $h_3$] from the outputs of the two local pathways $\mathcal{S}_1$, $\mathcal{S}_2$ and the global pathway $\mathcal{S}_g$ for constructing the Laplacian Pyramid. Since the local pathway $\mathcal{S}_3$ performs restoration last in all four pathways, we use its restored background image as the low-frequency component $l$. By learning a mask $\mathbf{M}$ for removing the noise constituents of the high-frequency components, our Laplacian Pyramid can obtain noise-free high-frequency components:
\begin{equation}
    h'_i = h_i \odot \mathbf{M}_i,
\end{equation}
where $h_i$ is the high-frequency components in the $i$-th pathway.
Finally, according to the reversibility of Laplacian Pyramid, we can synthesize the final restored image $\hat{\mathbf{B}}_{\text{out}}$ by combining the high-frequency components [$h'_1, h'_2, h'_3$] and low-frequency component $l$:
\begin{equation}
    \hat{\mathbf{B}}_{\text{out}} = h'_1 + \text{Up}(h'_2+\text{Up}(h'_3+l)),
\end{equation}
where $\text{Up}$ denotes bi-linear interpolation upsampling.
\subsection{Jointly Supervised Parameter Learning}
We optimize the parameters of our \emph{GLSGN} in an end-to-end manner.

\smallskip\noindent\textbf{Multi-stage Pixel Reconstruction Loss.} We employ the $L_1$ loss to learn the pixel level reconstruction of each pathway:
\begin{small}
\begin{equation}
    \mathcal{L}_{\text{pixel}}=\alpha_1\sum_{i=1}^{N}{\mathcal{L}_1\left(\mathbf{B},{\hat{\mathbf{B}}}_i\right)} + \beta_1\mathcal{L}_1\left(\mathbf{B},{\hat{\mathbf{B}}}_{\text{out}}\right),
\end{equation}
\end{small}\noindent 
where ${\hat{\mathbf{B}}}_i$ denotes the restored image in the $i$-th pathway, and ${\hat{\mathbf{B}}}_{\text{out}}$ is the restored image of Laplacian Pyramid. Besides, we obtain groundtruth $\mathbf{B}$ of various scales by interpolation. Empirically, we set $\alpha_1\!=\!0.1$ and $\beta_1=0.5$.

\smallskip\noindent\textbf{Multi-stage Perceptual Loss.} To learn the consistency of semantic information, we perform supervision of the perceptual loss~\cite{johnson2016perceptual} in each pathway:
\begin{small}
\begin{equation}
    \mathcal{L}_\text{perc}=\alpha_2\sum_{i=1}^{N}{\mathcal{L}_{\text{VGG}}\left(\mathbf{B},{\hat{\mathbf{B}}}_i\right)} + \beta_2\mathcal{L}_{\text{VGG}}\left(\mathbf{B},{\hat{\mathbf{B}}}_{\text{out}}\right),
\end{equation}
\end{small} 

\noindent where $\mathcal{L}_{\text{VGG}}$ denotes perceptual distance between two images measured by a pre-trained VGG-19~\cite{simonyan2014very}. Empirically, we set $\alpha_2\!=\!0.1$ and $\beta_2\!=\!0.5$. 

\smallskip\noindent \textbf{Conditional Adversarial Loss}, which encourages the restored image $\hat{\mathbf{B}}_{\text{out}}$ to be as realistic as the ground-truth background image $\mathbf{B}$. We employ the spectral normalization~\cite{miyato2018spectral} in our discriminator to stabilize the adversarial learning process:
\begin{equation}
    \mathcal{L}_{\text{adv}}  = -\mathbb{E}_{\text{B}\backsim \mathbb{P}_\text{GLSGN}}[D^{sn}(G(\mathbf{I}))], 
\end{equation}
where $D^{sn}$ is the discriminator with the spectral normalization module after each convolution layer.
In sum, the total loss function is defined as follows:
\begin{equation}
    \mathcal{L} = \lambda_1\cdot\mathcal{L}_{\text{pixel}}+\lambda_2\cdot\mathcal{L}_{\text{perc}}+\lambda_3\cdot\mathcal{L}_{\text{adv}}.
\end{equation}
We set the hyper-parameters $\lambda_1=1$ and $\lambda_2=\lambda_3=0.01$ to balance between different loss functions.

%% file: 4.0.dataset.tex
We construct the first ultra high-resolution (4K) benchmark dataset, UHR4K, for both reflection removal and rain streak removal.
UHR4K comprises 4,670 real-world and synthetic images, which are grouped into three subsets: UHR4K-Real and UHR4K-Syn for reflection removal, and UHR4K-Rain for rain streak removal. Figure~\ref{fig5:dataset} presents several randomly selected samples of three subsets.

\subsection{Image Reflection Removal}
\noindent\textbf{UHR4K-Real.} 
We collect real-world 4K images using two cameras, Nikon D300 and Huawei mobile phone.
We follow two steps to collect paired background and degraded images:
1) The reflection-contaminated image is first captured through the glass; 2) The groundtruth background image is captured by removing the glass.
Altogether 336 image pairs are captured, and we divide them into 316 image pairs as training set and 20 image pairs as test set.
To ensure that the model learns the reflection patterns rather than the image content, the real images from diverse scenes are collected. We also consider various reflection conditions while photographing real-world images including 1) circumstance: indoor and outdoor; 2) illumination: lamplight and sunlight; 3) glass thickness: 3mm and 9mm; 4) camera exposure value: 8.0-16.0; 5) camera view: front view and oblique view; 6) camera aperture: f/4.0 and f/16.0.

Due to the refraction effect of the glass, the inevitable spatial shifts still exist between the blended image taken behind the glass and the background image without the glass. To align the images as closely as possible, like Wan \emph{et al}.~\cite{wan2017benchmarking}, we first extract the ORB feature points~\cite{rublee2011orb} from the two images. Then the RANSAC algorithm~\cite{fischler1981random} is employed to calculate the homographic transformation matrix. Finally, the blended image is aligned to the background image by using the estimated transformation.
\begin{figure*}[!t]
    \centering
    \begin{minipage}[b]{1\linewidth}
    \includegraphics[width=\linewidth]{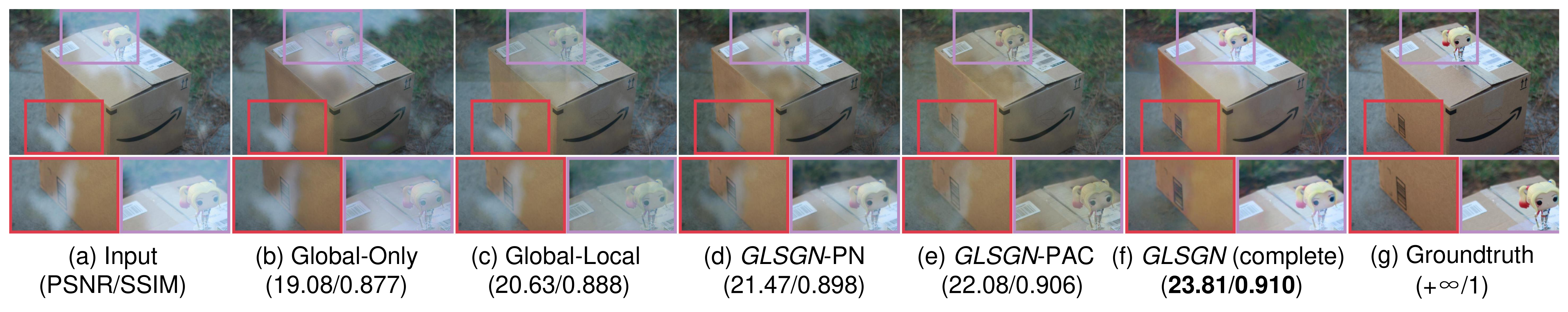}\vspace{4pt}
    \end{minipage}
    \caption{Reflection removal results by five variants of \emph{GLSGN} on a test image from Real20~\cite{zhang2018single} dataset.}
    \label{fig7:ablation}
\end{figure*}
\input{table5}

\noindent\textbf{UHR4K-Syn.} 
To further expand the reflection scenes, we extract 2,167 frames from numerous 4K videos as background, covering a wide variety of scenes.
Following the previous approach~\cite{fan2017generic}, we synthesize the reflection image by applying the Gaussian kernel to blur the images of DIV2K dataset~\cite{agustsson2017ntire}. 
Finally, the reflection image and background image are combined to synthesize the degraded image according to the physical principle~\cite{fan2017generic} of reflection formation.
Consequently, we obtain 2,167 image pairs, and divide 2,117 image pairs as training set and 50 image pairs as test set.


\begin{table}[!t]
\centering
\caption{Comparison with related real-world datasets for image reflection removal. The statistics are based on the image resolution, image amount, and scenes.}
\renewcommand\arraystretch{1.1}
\resizebox{0.875\linewidth}{!}{
\begin{tabular}{l|cccc}
\toprule
Dataset&Resolution & Amount & Scenes &  \\
\midrule
Solid~\cite{wan2017benchmarking}&540$\times$400&200&20 \\
Postcard~\cite{wan2017benchmarking}&540$\times$400&200&5  \\
Wild~\cite{wan2017benchmarking}&540$\times$400&55&50 \\
Nature~\cite{li2020single}&600$\times$400&220&67 \\
Real20~\cite{zhang2018single}&$\approx$1106$\times$902&110&83 \\
UHR4K-Real&\textbf{3840$\times$2160}&\textbf{336}&\textbf{116} \\
\bottomrule
\end{tabular}}
\label{tab:eff}
\end{table}

\subsection{Image Rain Streak Removal}
\noindent\textbf{UHR4K-Rain.} Due to the difficulty of capturing real-world image pairs for image rain streak removal, we simulate the rain conditions by employing the photoshop tutorial~\cite{RN16} to synthesize rain streaks. 
First, we randomly generate noise points with different densities. Then, we simulate rain streaks of different sizes and directions by stretching and rotating the direction of the noise with the Gaussian blur kernel. Finally, we superimpose the generated rain streaks mask onto the background image for the synthetic rain image. 
To ensure diversity, the generated rain streak images have different orientations, densities, sizes, and contrasts. Similar to the synthetic reflection dataset, we synthesize 2,167 rainy images using the background images we collected. Among them, 2,117 image pairs are used for training and 50 are used for testing.

\subsection{Comparison with Existing Datasets}
Compared to existing datasets~\cite{wan2017benchmarking, zhang2018single, li2020single, yang2017deep, zhang2018density} for image reflection removal and rain streak removal, our UHR4K enjoys the following advantages:
1) Much higher resolution (3840$\times$2160) of images, which challenges most frameworks for image restoration; 2) containing both synthetic and real-world images, which enables more comprehensive evaluation; 3) various degradations considered to make the dataset more realistic.

There is no real-world dataset and only one synthetic dataset~\cite{zheng2021ultra} for 4K image restoration to date, thus we construct UHR4K by collecting both synthetic and real-world images.
Compared with popular real-world datasets for reflection removal in Table~\ref{tab:eff}, our dataset contains more real-world images covering more scenes.
We refer to Real20~\cite{li2020single}
and Nature~\cite{zhang2018single} to set the split ratio of training and test set and amount of test images.
Our dataset has no overlapped scenes between training and test set while the other two datasets do.
We randomly select 20 images as held-out validation set to guide the model training.

%% file: table5.tex
\begin{table*}[!t]
\centering
\caption{Model complexity of our \emph{GLSGN} and six state-of-the-art methods for image restoration on a 4K test image in terms of TFLOPS, trainable parameters and runtime.}
\renewcommand\arraystretch{1.1}
\begin{tabular}{l|ccccccc}
    \toprule
    Methods & RmNet~\cite{wen2019single} & ERRNet~\cite{wei2019single} & IBCLN~\cite{li2020single} & MSBDN~\cite{dong2020multi}  & DMGN~\cite{feng2021deep}  & Zheng \emph{et al}.~\cite{zheng2021ultra} & \textbf{Ours} \\
    \midrule
    TFLOPS & 6.785 & 26.85 & 15.63 & 5.253  & 6.045 & \textbf{0.182} & 0.625 \\
    Params(M) & 65.43 & 18.95 & 21.61 & 31.35  & 45.49 & 34.54 & \textbf{15.69} \\
    Runtime(s) & 0.407 & 3.210 & 0.682 & 0.875  & 0.767 & 0.464 & \textbf{0.082} \\
    \bottomrule
    \end{tabular}%
\label{tab:efficiency}
\end{table*} 

%% file: 4.experiments.tex
\subsection{Implementation Details}
We implement our \emph{GLSGN} in distribution mode with 4 RTX 3090 GPUs under Pytorch, and employ the Adam~\cite{kingma2014adam} to optimize the process of gradient descent with batch-size 4. The learning rate is 2$\times10^{-4}$ during 500 training epochs. In our experiments, we randomly crop training images into 960$\times$540, and augment training by random flipping and rotation. 
In addition, we retrain all methods according to their released official codes and papers for fair comparison on ultra high-resolution images.




\begin{figure}[!t]
    \begin{center}
    \centering
    \begin{minipage}[b]{0.9\linewidth}
    \includegraphics[width=\linewidth]{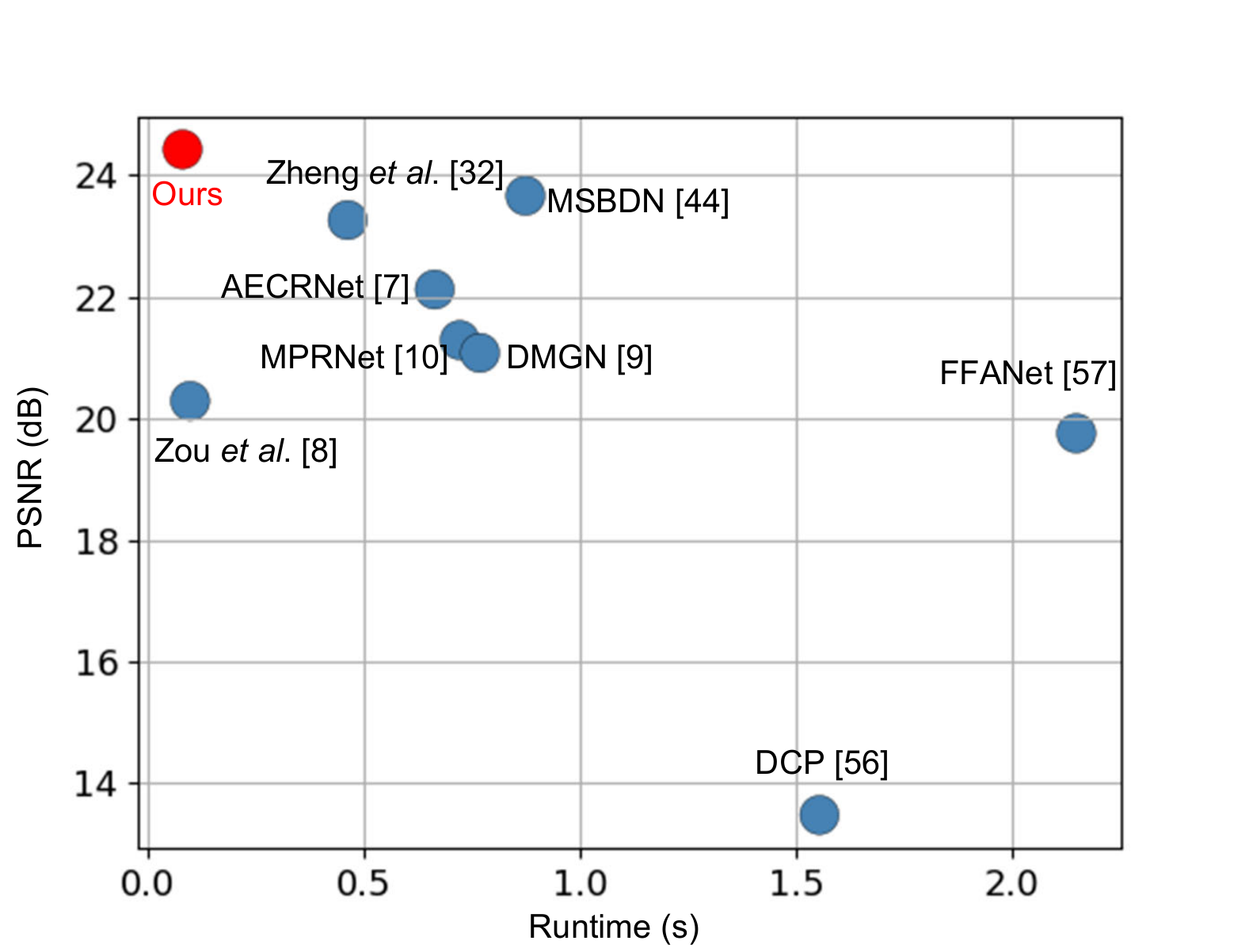}
    \end{minipage}
    \caption{Runtime vs Performance.}
    \label{fig8:runtime}
    \end{center}
\end{figure}

\input{table1}

\subsection{Ablation Study}
\noindent\textbf{Ablation experiments on reflection removal.} To investigate functional technique components in our \emph{GLSGN}, we conduct ablation study on the reflection removal task,
and perform experiments on five variants of our \emph{GLSGN}:
1) \textbf{Global-Only}, which only employs the global pathway; 
2) \textbf{Global-Local}, which collaboratively employs both the global pathway and local pathways; 
3) \textbf{\emph{GLSGN}-PN} that brings in the patch-normalization; 
4) \textbf{\emph{GLSGN}-PAC} that leverages the proposed perceptual attention consistency; 
5) \textbf{\emph{GLSGN}(complete)} further employs the Laplacian Pyramid for image reconstruction.

Table~\ref{tab1:ablation} presents the experimental results of ablation study. The comparison between \textbf{Global-Only} and \textbf{Global-Local} demonstrates the necessity of global-local collaboration. The increasingly better quantitative performance of five variants shows the effectiveness of proposed technical components. Furthermore, we visualize the restored results of various variants in Figure~\ref{fig7:ablation} on a randomly selected test image, which also illustrates the superiority of each component in our \emph{GLSGN} for thoroughly removing noise patterns. 

\begin{figure}[!t]
    \centering
    \begin{minipage}[b]{1\linewidth}
    \includegraphics[width=\linewidth]{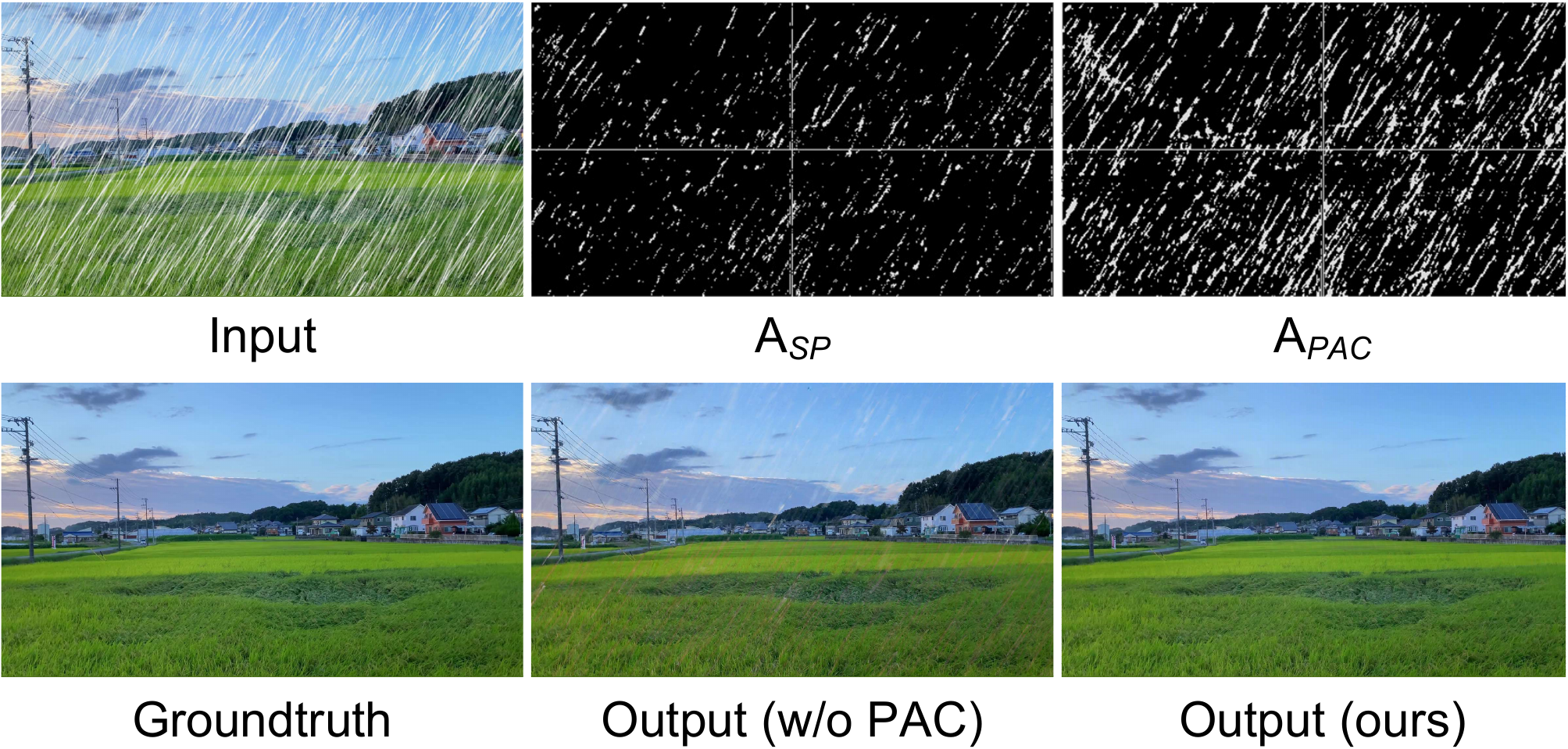}
    \end{minipage}
    \caption{Visualization of rain streak perception in the second local pathway on a randomly selected 4K test image for rain streak removal. A$_{S\!P}$ is the attention maps of plain spatial attention, and A$_{P\!A\!C}$ is the attention maps of PAC mechanism.
    }
    \label{fig7:PCA}
    \vspace{-4pt}
\end{figure}
\begin{figure}[!t]
    \centering
    \begin{minipage}[b]{0.925\linewidth}
    \includegraphics[width=\linewidth]{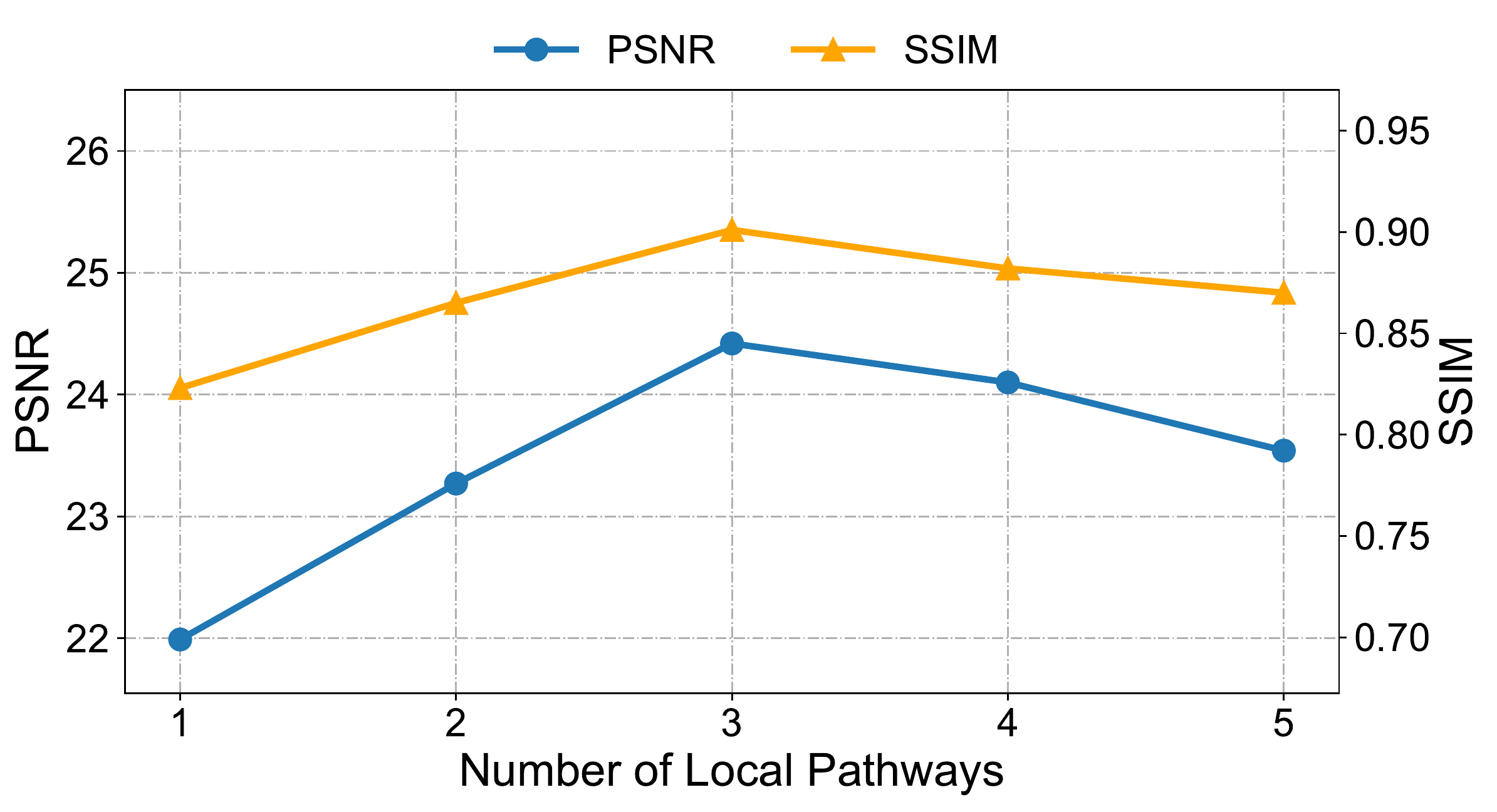}
    \end{minipage}
    \caption{Experimental results of different numbers of local pathways in \emph{GLSGN} on 4KID~\cite{zheng2021ultra} dataset. The local pathway number N=3 yields the best performance in terms of PSNR and SSIM. 
    }
    \label{fig8:stage}
\end{figure}

\noindent\textbf{Effect of the perceptual attention consistency}. Figure~\ref{fig7:PCA} visualizes the attention maps of our perceptual attention consistency and plain spatial attention~\cite{feng2021deep}. Our cross-pathway perceptual attention consistency obviously enables the framework to model rain streaks more completely and synthesize higher quality results than plain spatial attention. 


\begin{table}[!t]
\centering
\caption{Runtime and memory usage on different devices.}
\renewcommand\arraystretch{1.1}
\resizebox{1\linewidth}{!}{
\Huge
\begin{tabular}{l|ccccc|cc}
\toprule
Method &\multicolumn{5}{c|}{GLSGN (ours)}&MPRNet~\cite{zamir2021multi}&DMGN~\cite{feng2021deep} \\

\cmidrule(lr){0-1}
\cmidrule(lr){2-6}
\cmidrule(lr){7-8}

Device &RTX3090&RTX Titan& RTX 2080 &Titan xp &CPU &\multicolumn{2}{c}{RTX3090} \\

\midrule
Runtime(s)&0.082 &0.113&0.142&0.155&3.980 &0.549&0.767 \\
Memory(GB) &\multicolumn{5}{c|}{8.626}&17.72&21.60 \\
\bottomrule
\end{tabular}}
\label{tab:devices}
\end{table}

\noindent\textbf{Investigation on the model efficiency.}
To investigate whether the performance superiority of our \emph{GLSGN} over competing methods is benefited from the advantages of model design or model complexity, we compare the model efficiency of \emph{GLSGN} with six state-of-the-art methods~\cite{wen2019single, wei2019single, li2020single, dong2020multi, feng2021deep, zheng2021ultra} for image restoration in Table~\ref{tab:efficiency}. 
Besides, as shown in Figure~\ref{fig8:runtime}, we also perform the comparison between \emph{GLSGN} and other methods on 4KID~\cite{zheng2021ultra} dataset in terms of model performance and runtime on the dehazing task. 
It demonstrates our \emph{GLSGN} not only enjoys less TFLOPS, trainable parameters and runtime, but also achieves the best performance than most competing methods.
To further investigate the efficiency of our proposed \emph{GLSGN} on a 4K test image, we list the runtime and memory usage on various computational devices in Table~\ref{tab:devices}. The results demonstrate the high efficiency of our proposed \emph{GLSGN}.

\noindent\textbf{Investigation on the number of local pathways.} To explore how many pathways are optimal to predict the clean background image, we train the model with different numbers of local pathways on ultra high-resolution image dehazing. 
The experimental results are illustrated in Figure~\ref{fig8:stage}, which shows deploying 3 local pathways achieves the best performance.

\begin{figure*}[!t]
    \centering
    \begin{minipage}[b]{1\linewidth}
    \includegraphics[width=\linewidth]{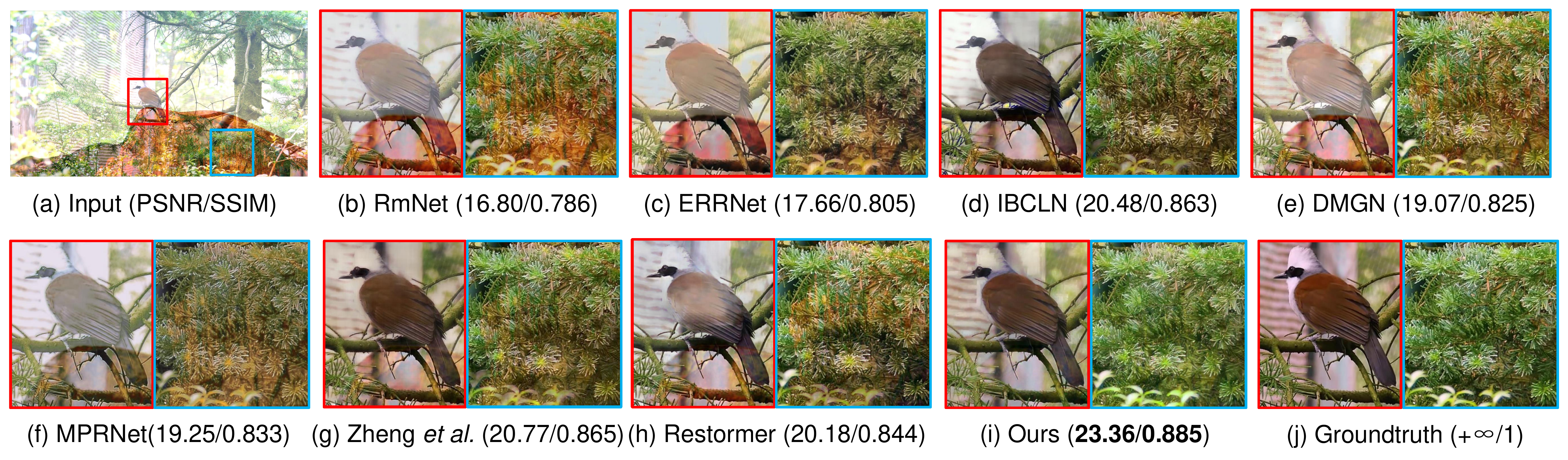}
    \end{minipage}
    \vspace{-16pt}
    \caption{Visualization of restored images for reflection removal on UHR4K-Syn dataset. 
    }
    \label{fig7:syn}
\end{figure*}
\begin{figure*}[!t]
    \centering
    \begin{minipage}[b]{1\linewidth}
    \includegraphics[width=\linewidth]{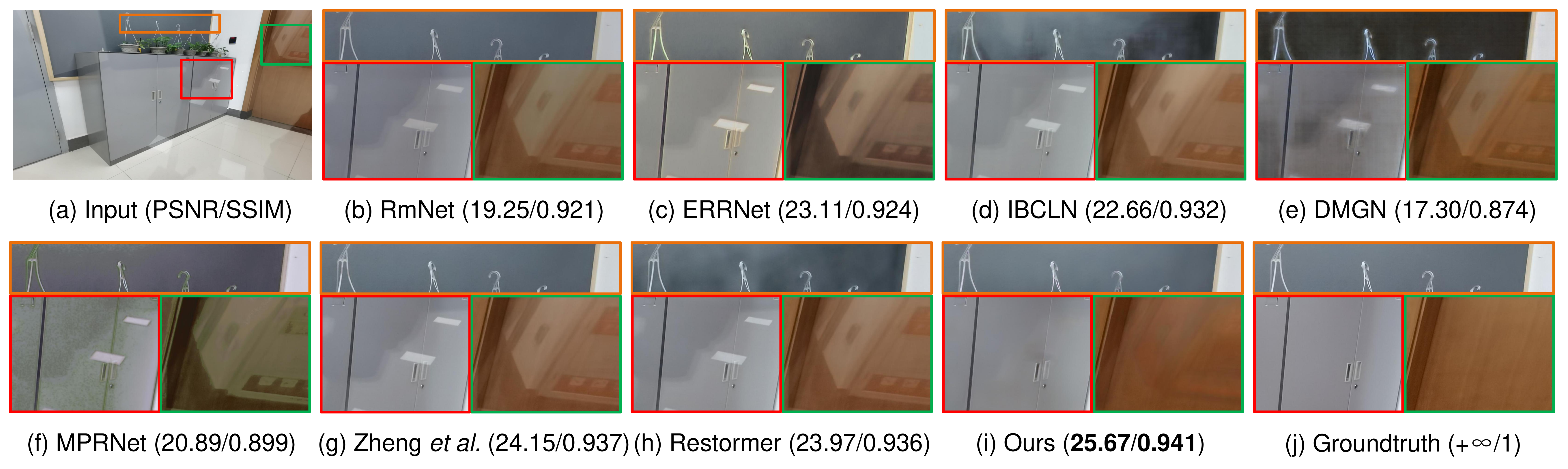}
    \end{minipage}
    \vspace{-16pt}
    \caption{
    Visualization of restored images for reflection removal on UHR4K-Real dataset.
    }
    \label{fig7:real}
\end{figure*}

\begin{table}[!t]
    \centering
    \caption{Analysis of SSIM loss and high-frequency scales.}
    \renewcommand\arraystretch{1.1}
    \resizebox{1\linewidth}{!}{
        \large
        \begin{tabular}{l|c|cccc}
    \toprule
    \multicolumn{1}{l|}{Dataset} &Metric& GLSGN$_{\text{SSIM}}$ &GLSGN$_{h_0}$&GLSGN$_{h_0+h_1}$ & Ours \\
    \midrule
    \multirow{2}{*}{UHR4K-Real}&PSNR&23.54&23.58&23.86&\textbf{24.35}\\
    ~&SSIM&0.832&0.831&0.836&\textbf{0.841}\\
    \bottomrule
    \end{tabular}%
    }
    \label{tab1:ssim}
\end{table}
\noindent\textbf{Investigation on Laplacian pyramid (LP).}
We conduct reflection removal experiments on UHR4K-Real dataset to investigate 1) the difference between LP and SSIM loss, 2) the effect of different scale of high-frequency information in LP, which is shown in Table~\ref{tab1:ssim}. The results show that LP performs better than SSIM loss distinctly and it is important to incorporate different scales of high-frequency information.
\begin{table}[!t]
    \centering
    \caption{Human evaluation on the reflection removal results. 50 human subjects are asked to perform comparison between our model and other three methods on the restored background images of 50 randomly selected test samples. Our model obtains $75.2\%$ votes among $50\times 50 = 2500$ comparisons and wins on 44 samples.}
    \renewcommand\arraystretch{1.1}
    \resizebox{0.8\linewidth}{!}{
    \begin{tabular}{l|cc}
    \toprule
    Model  &Share of the vote &Winning samples \\
    \midrule
ERRNet~\cite{wei2019single} & 1.8$\%$  &0 \\
IBCLN~\cite{li2020single}  &10.6$\%$   &2 \\
DMGN~\cite{feng2021deep}   & 12.4$\%$    &4 \\
Ours &75.2$\%$ &44 \\
\bottomrule
    \end{tabular}
    }
    \label{tab5_1:human_evalution}
\end{table}

\noindent\textbf{User study.} This section further presents the results of user study because PSNR and SSIM have their bias for quality evaluation of restored background images. We perform human evaluation
to compare our model with other top-three most powerful methods for reflection removal including ERRNet~\cite{wei2019single}, IBCLN~\cite{li2020single} and DMGN~\cite{feng2021deep}. We randomly select 50 test samples and present the generated background images by our model and other three methods to 50 human subjects for manual comparison of restoring quality. As shown in Table~\ref{tab5_1:human_evalution}, among total 50×50 = 2500 comparisons, our model wins on 75.2$\%$ of 2,500 votes which is much more than other methods. Besides, we aggregate evaluation results of all subjects for each sample, and our model wins on 44 test samples and fails on only 6 test samples.
\begin{figure*}[htbp]
    \centering
    \begin{minipage}[b]{1\linewidth}
    \includegraphics[width=\linewidth]{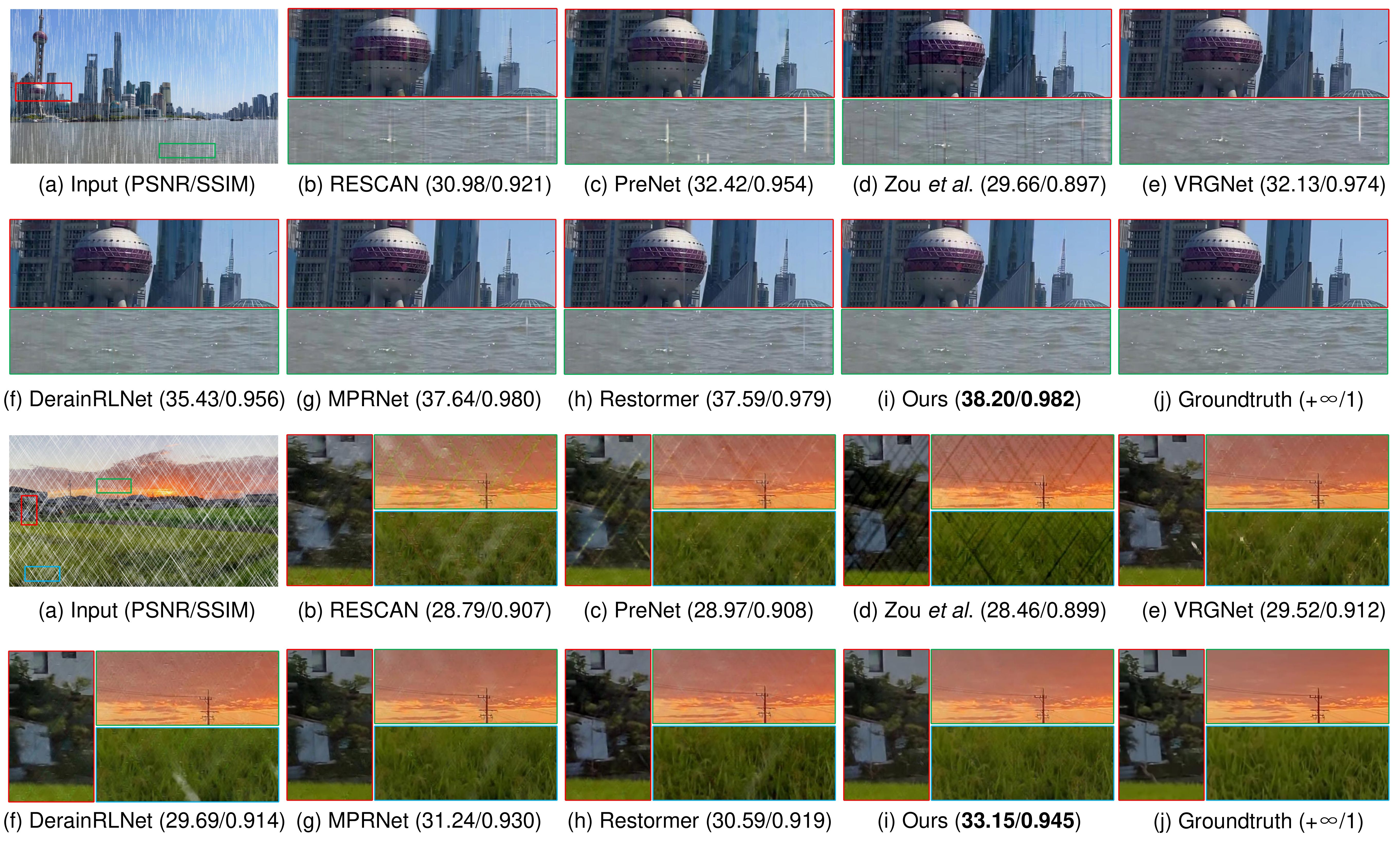}\vspace{4pt}
    \end{minipage}
    \caption{Visualization of restored images for rain streak removal on UHR4K-Rain dataset. 
    }
    \label{fig8:derain}
\end{figure*}

\input{table2}

\subsection{Comparison with State-of-the-art Methods}
\subsubsection{\textbf{Experiments on image reflection removal}}
We conduct experiments to compare our \emph{GLSGN} with nine state-of-the-art methods~\cite{zou2020deep,feng2021deep,zamir2021multi,wen2019single,zheng2021ultra,wei2019single,li2020single,feng2021contrastive,Zamir2021Restormer} for image reflection removal on five datasets including our proposed ultra high-resolution datasets as UHR4K-Real and UHR4K-Syn, and multiple public benchmark datasets with regular resolution as Real20~\cite{zhang2018single}, Nature~\cite{li2020single}, and $\text{SIR}^2$~\cite{wan2017benchmarking}. Table~\ref{tab3:reflection} lists the quantitative results of various methods for image reflection removal on five datasets in terms of PSNR and SSIM. Our \emph{GLSGN} achieves significant improvement on ultra high-resolution datasets compared to other competing methods. Meanwhile, on several public benchmark datasets, our method also obtains comparable results against state-of-the-art methods. It is reasonable that stronger capability of noise modeling drives our model to enjoy higher performance and thus \emph{GLSGN} is able to restore higher-quality image from the degraded image. The large gap between Zheng \emph{et al.}~\cite{zheng2021ultra} and our \emph{GLSGN} demonstrates that our model has significant superiority for restoring ultra high-resolution images over the method specifically designed for high-resolution image restoration, even on challenging reflection removal task.

To gain more insights into the model performance of various methods, we randomly select two examples from the synthetic and real-world test sets and visualize their restored images in Figure~\ref{fig7:syn} and Figure~\ref{fig7:real} respectively. Our \emph{GLSGN} is able to remove most of reflection in the input image and synthesize cleaner background images. 

\begin{figure*}[htbp]
    \centering
    \begin{minipage}[b]{1\linewidth}
    \includegraphics[width=\linewidth]{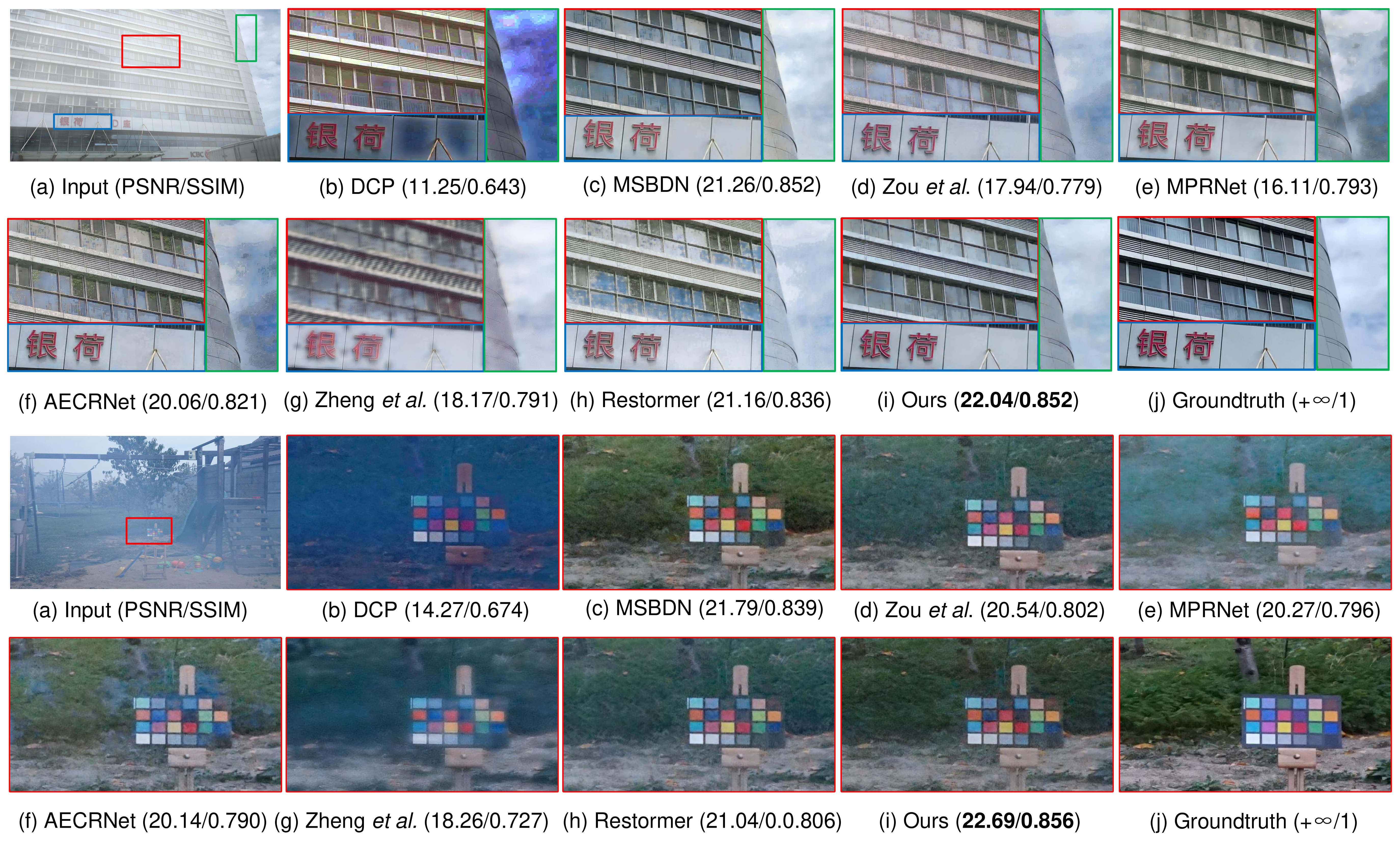}\vspace{4pt}
    \end{minipage}
    \caption{Visualization of restored images for image dehazing on dehazing dataset. 
    }
    \label{fig7:dehaze}
\end{figure*}
\input{table3}

\subsubsection{\textbf{Experiments on image rain streak removal}}
We conduct experiments to compare our \emph{GLSGN} with nine state-of-the-art methods~\cite{li2018recurrent,ren2019progressive,chen2021robust,wang2021rain,zou2020deep,feng2021deep,zamir2021multi,zheng2021ultra,Zamir2021Restormer} for image rain streak removal on our collected UHR4K-Rain dataset and two public benchmark datasets with regular resolution: Rain100H~\cite{yang2017deep} and Rain100L~\cite{yang2017deep} datasets.
Table~\ref{tab4:derain} presents the quantitative results of various methods for rain streak removal, and our \emph{GLSGN} consistently achieves the best performance on UHR4K-Rain and Rain100L datasets over nine methods. In particular, on UHR4K-Rain dataset, our method achieves significant improvement over other methods, indicating the strength of our model in processing ultra high-resolution images. On Rain100H dataset, our method obtains 
similar performance compared to the state-of-the-art methods for image rain streak removal. The experimental results indicate the superiority of our \emph{GLSGN} on both regular resolution and high-resolution images.


To have a qualitative comparison, we visualize the restored images by different methods for image rain streak removal on two randomly selected examples from the test set in Figure~\ref{fig8:derain}. In particular, our \emph{GLSGN} performs distinctly better than other competing methods for rain streak removal.
Benefiting from effective collaboration between various pathways, our \emph{GLSGN} can more thoroughly remove rain streaks in ultra high-resolution images.


\subsubsection{\textbf{Experiments on image dehazing}}
We conduct ultra high-resolution image dehazing experiments on existing datasets OHaze~\cite{ancuti2018haze} and 4KID~\cite{zheng2021ultra} to compare our proposed \emph{GLSGN} with nine current state-of-the-art methods~\cite{he2010single,qin2020ffa,dong2020multi,zou2020deep,feng2021deep,wu2021contrastive,zamir2021multi,zheng2021ultra,Zamir2021Restormer}.
Table~\ref{tab:dehaze} presents the quantitative results of different methods for image dehazing in terms of PSNR and SSIM, and our \emph{GLSGN} surpasses all other state-of-the-art models by a large margin on both OHaze~\cite{ancuti2018haze} and 4KID~\cite{zheng2021ultra} datasets.
\input{table4}

We visualize the restored details of the background image by different methods on two randomly selected test images in Figure~\ref{fig7:dehaze}.
It manifests that haze patterns are removed more thoroughly by our \emph{GLSGN} than other competing methods. Especially, compared to
Zheng \emph{et al}.~\cite{zheng2021ultra} which is a method specifically designed for ultra high-resolution image dehazing, 
our \emph{GLSGN} is able to synthesize higher quality background image due to the consideration of collaborative consistency among pathways.


%% file: table1.tex
\begin{table}[!t]
    \centering
    \caption{Ablation study on our \emph{GLSGN} in terms of PSNR and SSIM to investigate the effectiveness of each proposed technique in our model.}
    \vspace{-5pt}
    \renewcommand\arraystretch{1.1}
    \resizebox{0.875\linewidth}{!}{
    \begin{tabular}{l|cccc}
    \toprule
    \multirow{2}{*}{Method} & \multicolumn{2}{c}{Real20~\cite{zhang2018single}} & \multicolumn{2}{c}{UHR4K-Real}\\
    \cmidrule(lr){2-3}
    \cmidrule(lr){4-5}
    ~& PSNR & SSIM & PSNR & SSIM\\
    \midrule
    Global-Only & 19.42 & 0.712 & 20.54 & 0.785\\
    Global-Local & 20.37 & 0.739 & 21.88 & 0.801 \\
    \emph{GLSGN}-PN & 21.02 & 0.754 & 22.48 & 0.812 \\
    \emph{GLSGN}-PAC  & 21.65  & 0.778 & 23.46 & 0.830\\
    \emph{GLSGN} (complete) & \textbf{22.20} & \textbf{0.790} & \textbf{24.35} & \textbf{0.841}\\
    \bottomrule
    \end{tabular}
    }
    \label{tab1:ablation}
    \vspace{-4pt}
\end{table}

%% file: table2.tex
\begin{table*}[!t]
    \centering
    \caption{Quantitative results of different models for image reflection removal on five datasets in terms of PSNR and SSIM.}
    \renewcommand\arraystretch{1.1}
    \resizebox{0.725\linewidth}{!}{
    \large
    \begin{tabular}{l|ccccccccccc}
    \toprule
    \multirow{2}{*}{Method} & \multicolumn{2}{c}{Real20~\cite{zhang2018single}}& \multicolumn{2}{c}{Nature~\cite{li2020single}} &
    \multicolumn{2}{c}{SIR$^2$~\cite{wan2017benchmarking}} &\multicolumn{2}{c}{UHR4K-Syn} & \multicolumn{2}{c}{UHR4K-Real} \\
    \cmidrule(lr){2-3}
    \cmidrule(lr){4-5}
    \cmidrule(lr){6-7}
    \cmidrule(lr){8-9}
    \cmidrule(lr){10-11}
    & PSNR & SSIM& PSNR & SSIM & PSNR & SSIM & PSNR & SSIM& PSNR & SSIM  \\
    \midrule
    RmNet~\cite{wen2019single} & 19.47 & 0.748&19.07&0.755&20.74&0.836 & 17.05 & 0.829 & 20.86 & 0.808 \\
    ERRNet~\cite{wei2019single} & 18.87 & 0.735&20.79&0.796&22.60&0.856 & 17.54 & 0.818 & 21.69 & 0.819 \\
    IBCLN~\cite{li2020single} & 19.99 & 0.759&23.57&0.783&22.68&0.860 & 20.66 & 0.879 & 22.59 & 0.824 \\
    CFDNet~\cite{feng2021contrastive}  & 18.90  & 0.706 &23.79&0.811&23.89&0.884& 16.99 & 0.785 & 20.84 & 0.773 \\
    Zou \emph{et al}.~\cite{zou2020deep} & 19.76 & 0.752 &22.34&0.803&23.52&0.877& 18.89 & 0.841 & 20.57 & 0.799 \\
    DMGN~\cite{feng2021deep}  & 19.26 & 0.745 &23.23&0.839&23.41&0.875& 20.10 & 0.865 & 22.91 & 0.833 \\
    MPRNet~\cite{zamir2021multi} & 21.37 & 0.781 &23.42&0.848&23.82&0.880 & 21.26 & 0.888 & 22.71 & 0.821 \\
    Zheng \emph{et al}.~\cite{zheng2021ultra}&20.61& 0.756 &22.59&0.782&22.46&0.868&23.57&0.886 &20.08 &0.712\\
    Restormer~\cite{Zamir2021Restormer} &21.89 & 0.778  &23.98 &0.852 &\textbf{24.25} &0.889 &23.07 &0.882 &22.77 &0.822 \\
    \textbf{GLSGN (Ours)} & \textbf{22.20} & \textbf{0.790} &\textbf{24.27} &\textbf{0.856} &24.11 &\textbf{0.903} &\textbf{25.96} & \textbf{0.911} & \textbf{24.35} & \textbf{0.841} \\
\bottomrule
    \end{tabular}
    }
    \label{tab3:reflection}
\end{table*}

%% file: table3.tex
\begin{table}[!t]
    \centering
    \caption{Quantitative results of different models for rain streak removal on Rain100~\cite{yang2017deep} and UHR4K-Rain in terms of PSNR and SSIM.}
    \renewcommand\arraystretch{1.1}
    \resizebox{1\linewidth}{!}{
    \begin{tabular}{l|cccccc}
    \toprule
    \multirow{2}{*}{Method} & \multicolumn{2}{c}{Rain100L~\cite{yang2017deep}} & \multicolumn{2}{c}{Rain100H~\cite{yang2017deep}}  &
    \multicolumn{2}{c}{UHR4K-Rain}\\
    \cmidrule(lr){2-3}
    \cmidrule(lr){4-5}
    \cmidrule(lr){6-7}
    & PSNR & SSIM & PSNR & SSIM & PSNR & SSIM \\
    \midrule
RESCAN~\cite{li2018recurrent} &29.80&0.881&26.36&0.786&26.85&0.837 \\ 
PReNet~\cite{ren2019progressive} &32.44&0.950&26.77&0.858&27.08&0.865\\ 
DerainRLNet~\cite{chen2021robust} &37.49&0.968&30.38&0.899&31.46&0.920\\ 
VRGNet~\cite{wang2021rain} &37.21&0.970&30.82&0.904&28.81&0.913 \\
Zou \emph{et al}.~\cite{zou2020deep} &36.21&0.963&30.01&0.883&27.25&0.837\\ 
DMGN~\cite{feng2021deep}  &38.53&0.979&30.72&0.903&29.97&0.890\\ 
MPRNet~\cite{zamir2021multi} &36.40&0.965&30.41&0.890&31.64&0.929\\
Zheng \emph{et al}.~\cite{zheng2021ultra}&33.26&0.952&27.64&0.861&27.18&0.812\\
Restormer~\cite{Zamir2021Restormer} &38.99&0.978&\textbf{31.46}&\textbf{0.904}&30.93&0.917\\
\textbf{GLSGN (Ours)} &\textbf{39.12}&\textbf{0.980}&30.89&0.901&\textbf{32.83}&\textbf{0.938}\\
\bottomrule
    \end{tabular}
    }
    \label{tab4:derain}
    \vspace{-12pt}
\end{table}

%% file: table4.tex
\begin{table}[!t]
\vspace{2pt}
    \centering
    \caption{Quantitative results of different models for image dehazing on 4KID~\cite{zheng2021ultra} and OHaze~\cite{ancuti2018haze} in terms of PSNR and SSIM.}
    \renewcommand\arraystretch{1.1}
    \resizebox{0.875\linewidth}{!}{
    
    \begin{tabular}{l|cccc}
    \toprule
    \multirow{2}{*}{Method} & \multicolumn{2}{c}{4KID~\cite{zheng2021ultra}} & \multicolumn{2}{c}{OHaze~\cite{ancuti2018haze}} \\
    \cmidrule(lr){2-3}
    \cmidrule(lr){4-5}
    & PSNR & SSIM & PSNR & SSIM \\
    \midrule
DCP~\cite{he2010single}   & 13.48 & 0.739 & 14.86 & 0.601 \\
FFANet~\cite{qin2020ffa} & 19.76 & 0.869 & 22.55 & 0.754 \\
MSBDN~\cite{dong2020multi} & 23.06 & 0.872 & 22.72 & 0.751 \\
Zou \emph{et al.}~\cite{zou2020deep} & 20.29 & 0.855 & 22.16 & 0.742 \\
DMGN~\cite{feng2021deep}  & 21.08 & 0.866 & 20.54 & 0.734 \\
AECRNet~\cite{wu2021contrastive} & 22.12 & 0.872 & 23.16 & 0.762 \\
MPRNet~\cite{zamir2021multi} & 21.29 & 0.882 & 20.78 & 0.739 \\
Zheng \emph{et al}.~\cite{zheng2021ultra} & 23.43 & 0.843 & 21.41 & 0.693 \\
Restormer~\cite{Zamir2021Restormer} &22.41 &0.886 &21.71 &0.712 \\
\textbf{Ours} &\textbf{24.42}    &\textbf{0.901}   &\textbf{23.67}  &\textbf{0.768} \\

\bottomrule
    \end{tabular}
    }
    \label{tab:dehaze}
\end{table}

%% file: 5.conclusion.tex
In this work, we propose a Global-Local Stepwise Generative Network \emph{GLSGN} for ultra high-resolution image restoration.
Our \emph{GLSGN} adopts both local and global pathways to restore images in a stepwise manner.
To facilitate the collaboration of various pathways, our \emph{GLSGN} models the consistency between different pathways in terms of four aspects: low-level content, perceptual attention, restoring intensity and high-level semantics.
Furthermore, we establish a large scale benchmark dataset UHR4K for ultra high-resolution image restoration, which is the first dataset for image reflection removal and rain streak removal.

Though it has been shown that our \emph{GLSGN} is able to deal with various degradations individually, it is still challenging
for our method to handle multiple degradations simultaneously for a same image. In future work, we foresee several research directions, including 1) image restoration for a mixture of multiple degradations via learning generalizable noise patterns; 2) refinement of modeling structure over our method to improve the efficiency.

